\documentclass[sigconf]{acmart}
\AtBeginDocument{%
  \providecommand\BibTeX{{%
    \normalfont B\kern-0.5em{\scshape i\kern-0.25em b}\kern-0.8em\TeX}}}

\usepackage{multirow}
\usepackage{float} 
\usepackage{amsmath}
\usepackage[ruled,linesnumbered]{algorithm2e}
\usepackage{graphicx}
\usepackage{subfigure}
\usepackage{epstopdf}
\usepackage{tablefootnote}
\usepackage[flushleft]{threeparttable}
\usepackage{diagbox}
\usepackage{booktabs}
\usepackage{siunitx}
\usepackage{footmisc}
\usepackage{footnote}
\usepackage{enumitem}
\usepackage[normalem]{ulem}
\useunder{\uline}{\ul}{}

\copyrightyear{2023}
\acmYear{2023}
\setcopyright{rightsretained}\acmConference[COLIEE 2023]{COLIEE 2023 workshop: Competition on Legal Information Extraction/Entailment}{June 19, 2023}{Braga, Portugal}
\acmBooktitle{Proceedings of COLIEE 2023 workshop, June 19, 2023, Braga, Portugal}
\acmPrice{}
\acmISBN{}
\acmDOI{}

\begin{document}

\title{THUIR@COLIEE 2023: More Parameters and Legal Knowledge for Legal Case Entailment}

\author{Haitao Li}
\affiliation{DCST, Tsinghua University}
\affiliation{Quan Cheng Laboratory}
\affiliation{Beijing 100084, China}
\email{liht22@mails.tsinghua.edu.cn}

\author{Changyue Wang}
\affiliation{DCST, Tsinghua University}
\affiliation{Quan Cheng Laboratory}
\affiliation{Beijing 100084, China}
\email{changyue20@mails.tsinghua.edu.cn}

\author{Weihang Su}
\affiliation{DCST, Tsinghua University}
\affiliation{Quan Cheng Laboratory}
\affiliation{Beijing 100084, China}
\email{swh22@mails.tsinghua.edu.cn}

\author{Yueyue Wu}
\affiliation{DCST, Tsinghua University}
\affiliation{Quan Cheng Laboratory}
\affiliation{Beijing 100084, China}
\email{wuyueyue@mail.tsinghua.edu.cn}

\author{Qingyao Ai}
\affiliation{DCST, Tsinghua University}
\affiliation{Quan Cheng Laboratory}
\affiliation{Beijing 100084, China}
\email{aiqy@tsinghua.edu.cn}

\author{Yiqun Liu}
\authornote{Corresponding author}
\affiliation{DCST, Tsinghua University}
\affiliation{Quan Cheng Laboratory}
\affiliation{Beijing 100084, China}
\email{yiqunliu@tsinghua.edu.cn}

\begin{abstract}
This paper describes the approach of the THUIR team at the COLIEE 2023 Legal Case Entailment task. This task requires the participant to identify a specific paragraph from a given supporting case that entails the decision for the query case. We try traditional lexical matching methods and pre-trained language models with different sizes. Furthermore, learning-to-rank methods are employed to further improve performance. However, learning-to-rank is not very robust on this task. which suggests that answer passages cannot simply be determined with information retrieval techniques. Experimental results show that more parameters and legal knowledge contribute to the legal case entailment task. Finally, we get the third place in COLIEE 2023. The implementation of our method can be found at https://github.com/CSHaitao/THUIR-COLIEE2023.
\end{abstract}



\maketitle

\section{Introduction}
In countries with Case Law system, like the United States, Canada, etc, past precedent is an essential reference for making judicial judgments~\cite{locke2022case,shao2023understanding}. However, with the rapid growth of digital legal cases, legal practitioners need to expend significant effort to retrieve relevant documents and identify entailment parts. Recently, more and more researchers focus on intelligent legal systems to ease the heavy manual work~\cite{shao2020bert,bench2012history,yu2022explainable,ma2021lecard,althammer2021dossier,li2023sailer}.

As a well-known competition in the legal field, the Competition on Legal Information Extraction/Entailment (COLIEE) aims to achieve state-of-the-art methods to help the realization of intelligent legal systems~\cite{rabelo2022overview}. COLIEE contains two types of tasks: retrieval and entailment. The retrieval task is to identify the cases that support the query case from the large corpus~\cite{ma2021retrieving,nigam2023nigam,li2023constructing}. The entailment task identifies a specific paragraph from a given supporting case that entails the decision for the query case~\cite{rosa2021tune,rosa2022billions}. 

In this paper, we introduce the solution of the THUIR team for Legal Case Entailment task, which achieves third place and fifth place in the competition. To be specific, we formalize the entailment task as the paragraph ranking task. Then, we implemented several lexical matching models, such as BM25, QLD. Furthermore, contrastive learning loss is employed to fine-tune pre-trained models of different sizes. Finally, we utilize the above features to ensemble the final score. However, due to the sparse training data, the leaning to rank method does not achieve satisfactory performance, which may also indicate that the answer paragraphs cannot be simply confirmed by information retrieval techniques. As a result, THUIR teams placed third and fifth in 18 submissions from seven teams. Extensive experimental results show that more parameters and more legal knowledge contribute to better legal text understanding. The implementation of our method can be found at https://github.com/CSHaitao/THUIR-COLIEE2023.

\begin{figure*}[h]
\centering
\includegraphics[width=0.8\textwidth]{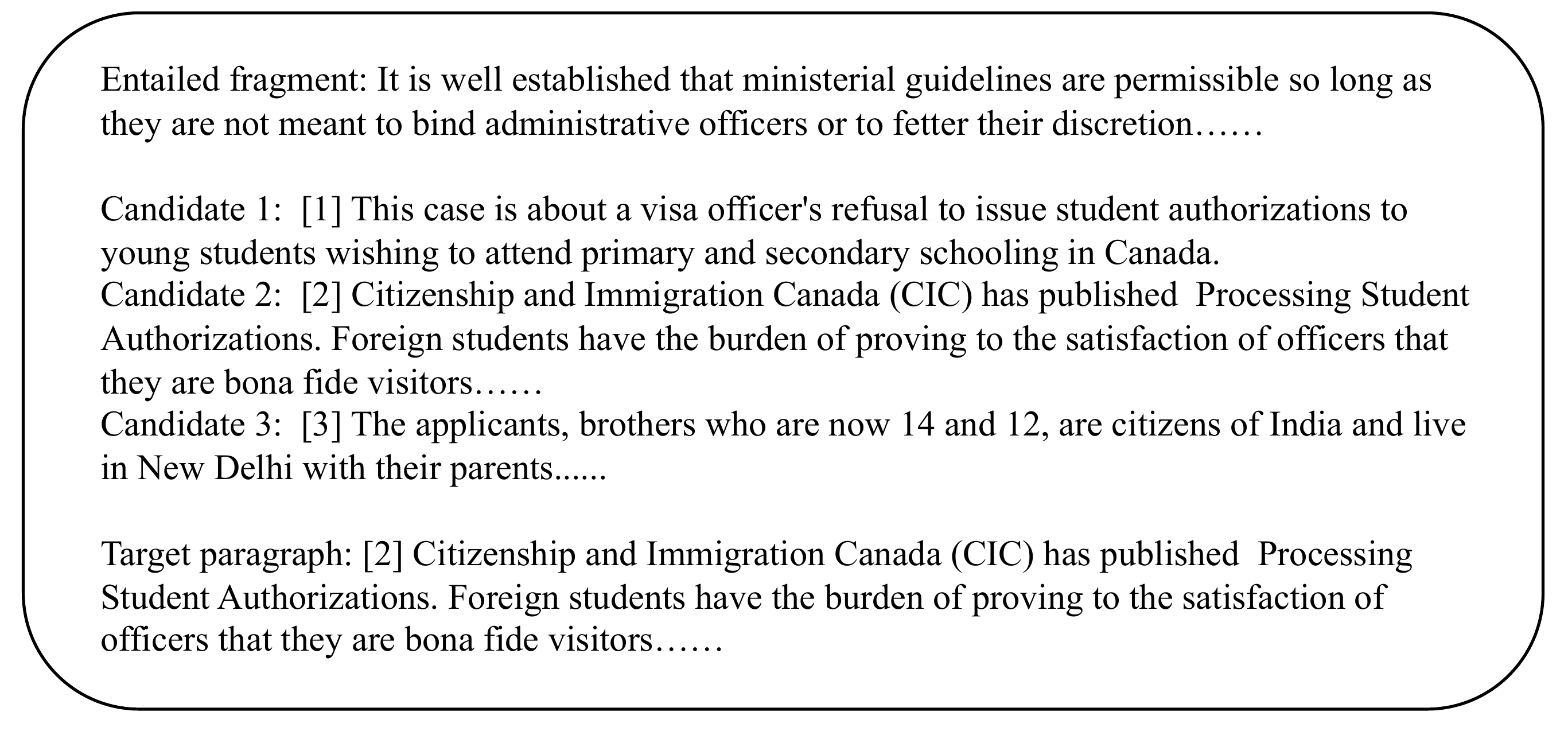}
\caption{An example of legal case entailment task.}
\label{example}
\end{figure*}

This paper is organized as follows: Section 2 introduces the related work. In Section 3, details of the legal case entailment task are elaborated. Section 4 shows our detailed methodology.
Then, the experimental setting and results are introduced in Section 5. Finally, we conclude our work and discuss the future direction in Section 6.

\section{Related Work}
Legal Case Entailment is an essential task in the legal field which aims to determine whether some specific paragraphs entail the decision of the query case. In past COLIEE competitions, various teams have achieved excellent results with traditional methods and deep learning methods, or a combination of both. For instance, the JNLP team~\cite{bui2022using} employs LEGAL-BERT and BM25 and try to capture keywords with ``Abstract Meaning Representation". NM team ~\cite{nigam2023nigam} explores the zero-sample learning potential of language models using monoT5, which wins the championship in COLIEE 2022. Furthermore, the TR group~\cite{schilder2021pentapus} employs hand-crafted similarity features and trains with random forest classifiers. The UA team~\cite{kim2021bm25}, on the other hand, utilizes methods such as fine-tuned language models and text summaries to accomplish the task. In this paper, we focus on the impact of more parameters and legal knowledge on legal case entailment task.

\section{Task Overview}
\subsection{Task Description}
The Legal case entailment task refers to identifying specific paragraphs from existing cases that entail the decision for the query case. Formally, given a query case $Q$ and a supporting case $R$ consisting of paragraphs $P = {P_1, P_2,..., P_n}$, this task is to determine the paragraph $P_i \in P$ that entails the decision for query case $Q$.

\subsection{Data Corpus}

The legal case entailment task is based on the existing collection of predominantly Federal Court of Canada case law. The training data includes the query case $Q$, a set of candidate paragraphs $P$ and the corresponding labels. Test data includes only query case $Q$, and candidate paragraphs $P$. Figure \ref{example} illustrates an example of legal case entailment task.

The COLIEE 2023 dataset contains 625 query cases for training and 100 query cases for testing. Table \ref{satictics} shows the statistics of datasets for the last three years. The dataset has an average of 35 candidate paragraphs for each query case, of which only one is relevant on average. We randomly selected 100 query cases as the validation set and the rest for training. Details of the validation set can be found on GitHub https://github.com/lihaitao18375278/THUIR-COLIEE2023.

\subsection{Metrics}
The evaluation metrics of legal case entailment task are precision, recall and
F1 score. Definition of these measures is as follows:

\begin{equation}
\text { Precision } = \frac{\# T P}{\# T P+\# F P} 
\end{equation}
\begin{equation}
\text { Recall } = \frac{\# T P}{\# T P+\# F N} 
\end{equation}
\begin{equation}
F-\text { measure } = \frac{2 \cdot \text { Precision } \cdot \text { Recall }}{\text { Precision }+ \text { Recall }}
\end{equation}

where $\#TP$ is the number of correctly retrieved candidate paragraphs for all query cases, $\#FP$ is the number of falsely retrieved candidate paragraphs for all query cases, and $\#FN$ is the number of missing noticed candidate paragraphs for all query cases.

\begin{table*}[ht]
\caption{Statistics of COLIEE Task 2.}
\begin{tabular}{l|cc|cc|cc}
\hline
                                      & \multicolumn{2}{c|}{2021} & \multicolumn{2}{c|}{2022} & \multicolumn{2}{c}{2023} \\
                                      & Train       & Test        & Train       & Test        & Train       & Test       \\ \hline
\# of query cases                     & 425         & 100         & 525         & 100         & 625         & 100        \\
Avg. \# of candidates per query       & 35.80       & 35.24       & 35.69       & 32.78       & 35.22       & 37.65      \\
Avg. \# positive candidates per query & 1.17        & 1.17        & 1.14        & 1.18        & 1.17        & 1.2        \\
Avg. query length                     & 37.51       & 32.97       & 36.64       & 32.21       & 35.36       & 36.57      \\
Avg. candidate length                 & 103.14      & 100.83      & 102.71      & 104.61      & 102.32      & 104.71     \\ \hline
\end{tabular}
\label{satictics}
\end{table*}

\begin{table*}[]
\caption{Features that we used for learning to rank.}
\begin{tabular}{cll}
\hline
\multicolumn{1}{l}{Feature ID} & Feature Name      & Description                                  \\ \hline
1                              & query\_length     & Length of the query                          \\
2                              & candidate\_length & Length of the candidate paragraph            \\
3                              & BM25              & Query-candidate scores with BM25             \\
4                              & QLD               & Query-candidate scores with QLD              \\
5                              & BERT-large        & Query-candidate scores with BERT-large       \\
6                              & RoBERTa-large     & Query-candidate scores with RoBERTa-large    \\
7                              & LEGAL-BERT-base    & Query-candidate scores with LEGAL-BERT-base   \\
8                              & DeBERTa-v3-large  & Query-candidate scores with DeBERTa-v3-large \\
9                              & monoT5-3B         & Query-candidate scores with monoT5-3B        \\ \hline
\end{tabular}
\label{feauture}
\end{table*}

\section{Method}
\subsection{Traditional Lexical Matching Model}
Traditional lexical matching models, i.e. BM25 and QLD, rank the candidate documents by a statistical probability model based on bag-of-words representation. In COLIEE 2023 task 2, we implement the following two lexical matching methods as baselines:

\begin{itemize}[leftmargin=*]
\item \textbf{BM25}~\cite{robertson2009probabilistic} is a classical lexical matching model with robust performance. The calculation formula of BM25 is shown in Eq ~\ref{eq:BM25 calculation}.
        \begin{equation}
	BM25(d, q) = \sum_{i = 1}^M \dfrac{IDF(t_i) \cdot TF(t_i, d)       \cdot (k_1+1)}{TF(t_i, d) + k_1 \cdot \left(1-b+b \cdot           \dfrac{len(d)}{avgdl}\right)}
	\label{eq:BM25 calculation}
        \end{equation}        
        where $k_1$, $b$ are hyperparameters, TF represents term frequency and IDF represents inverse document frequency.

\item \textbf{QLD}~\cite{zhai2008statistical} is another representative traditional retrieval model based on Dirichlet smoothing. The equation for QLD is shown in Eq ~\ref{eq:language model calculation}. 

    \begin{equation}
    	\log p(q|d) = \sum_{i: c(q_i; d)>0} \log \dfrac{p_s(q_i|d)}{\alpha_d p(q_i|\mathcal{C})} + n \log \alpha_d +\sum_i \log p(q_i|\mathcal{C})
    	\label{eq:language model calculation}
    \end{equation}
\end{itemize}

For better performance, we remove all placeholders in the paragraph, e.g. ``FRAGMANT\_SUPPRESSED", ``REFERENCE\_SUPPRESSED" etc. Pyserini toolkit~\footnote{\url{https://github.com/castorini/pyserini}} is employed to implement BM25 and QLD with default parameters.

\subsection{Pre-trained Language Model}

\subsubsection{Cross Encoder}

As pre-trained language models have shown great potential in legal case entailment task~\cite{dong2022incorporating,xie2023t2ranking}, we experimented with several pre-trained models of different sizes. More specifically, we employ the cross-encoder architecture. The scores of queries and candidates are defined as follows:

\begin{equation}
Input = [CLS] query [SEP] candidate [SEP]
 \end{equation}

\begin{equation}
 score(query,candidate) = MLP(CLS[PLM(Input)]) 
 \end{equation}
where CLS is the [CLS] token vector and MLP is a Multilayer Perceptron that projects the CLS vector to a score. PLM represents pre-trained language models. 
The purpose of training is to make the query case closer to related paragraphs in the vector space compared to the irrelevant ones. Thus, given a query case $q$, let $d^{+}$ and $d^{-}$ be relevant and negative paragraphs, the loss function $L$ is formulated as follows:

\begin{equation}\label{eqn-1} 
  L(q,d^+,d^-_{1},...,d^-_{n}) =
-\log_{}{    \frac{exp(s(q,d^+))}{exp(s(q,d^+))+\sum_{j=1}^nexp(s(q,d^-_j))}}
\end{equation}

We use the following model as the backbone for training:

\begin{itemize}[leftmargin=*]
\item \textbf{BERT}~\cite{devlin2018bert} is a multi-layer bidirectional Transformer encoder architecture that utilizes the Masked Language Model(MLM) and Next Sentence Prediction(NSP) as pre-training tasks.

\item \textbf{RoBERTa}~\cite{liu2019roberta} is an enhanced version of BERT with a more extensive dataset, which is pre-trained only with the Masked Language Model(MLM) task.

\item \textbf{LEGAL-BERT}~\cite{chalkidis2020legal} is pre-trained with extensive English legal database and has achieved state-of-the-art performance on multiple legal tasks

\item \textbf{DeBERTa}~\cite{he2020deberta} proposes a disentangled attention mechanism and an enhanced mask decoder to improve the original BERT architecture, achieving state-of-the-art performance in COLIEE task2 in previous years.
\end{itemize}

\subsubsection{Sequence-to-Sequence Model}

Applying cross encoder to the legal case entailment task can be seen as a classification-based approach. Correspondingly, sequence-to-sequence models have been widely explored in this task~\cite{rosa2022billions,rosa2021tune}. In general, sequence-to-sequence models perform better than cross encoders in data-poor settings due to capturing the underlying semantic relationships. In this section, we implement the following sequence-to-sequence models:

\begin{itemize}[leftmargin=*]
\item \textbf{monoT5}~\cite{nogueira2020document} is an encoder-decoder architecture. It generates ``true" or ``false" token based on the relevance of queries and candidates, and regards the probability of generating "true" as the final relevance score. To be specific, the input to monoT5 has the following form:
\begin{equation}
Input = Query: [Q] Document: [P] Relevant:
 \end{equation}
where [Q] and [P] are replaced with the query case and candidate paragraph texts, respectively. During the fine-tune process, ``true" and ``false" are the generated target token. At inference time, we calculate the probability of generating ``true" token to determine the final paragraphs ranking.

\item \textbf{FLAN-T5}~\cite{chung2022scaling} significantly improves zero-shot and few-shot abilities with instrution tuning and chain-of-thougt technology. FLAN-T5 is fine-tuned on 1836 tasks from 473 datasets with different instruction and improves zero-shot reasoning capability by a stepwise thinking approach.

\end{itemize}

\begin{table}[]
\caption{ Overall experimental results of different methods on COLIEE2023 validation set.}
\begin{tabular}{lcccc}
\hline
\textbf{Method}           & \multicolumn{1}{l}{\textbf{Params}} & \multicolumn{1}{l}{\textbf{P@1}} & \multicolumn{1}{l}{\textbf{R@1}} & \multicolumn{1}{l}{\textbf{F1 score}} \\ \hline
\multicolumn{5}{c}{\textbf{Lexical Matching Model}}                                                                              \\ \hline
BM25             & -                          & 0.670                   & 0.563                   & 0.612                        \\
QLD              & -                          & 0.632                   & 0.529                   & 0.575                        \\ \hline
\multicolumn{5}{c}{\textbf{Cross Encoder}}                                                                                       \\ \hline
BERT-base        & 110M                       & 0.750                   & 0.630                   & 0.685                        \\
BERT-large       & 340M                       & 0.780                   & 0.655                   & 0.712                        \\
RoBERTa-base     & 110M                       & 0.770                     & 0.648                        & 0.704                             \\
RoBERTa-large    & 340M                       & 0.810                   & 0.681                   & 0.739                        \\
LEGAL-BERT-base   & 110M                       & 0.830                   & 0.697                   & 0.758                        \\
DeBERTa-v3-base  & 110M                       & 0.790                   & 0.666                   & 0.721                        \\
DeBERTa-v3-large & 340M                       & 0.830                   & 0.697                   & 0.758                        \\ \hline
\multicolumn{5}{c}{\textbf{Sequence-to-Sequence Model}}                                                                          \\ \hline
monoT5-base      & 250M                       & 0.800                   & 0.672                   & 0.731                        \\
monoT5-3B        & 3000M                      & 0.850                   & 0.714                   & 0.776                        \\
FLAN-T5-base     & 250M                       & 0.700                   & 0.588                   & 0.639                        \\
FLAN-T5-3B       & 3000M                      & 0.730                   & 0.613                   & 0.666                        \\ \hline
Ensemble         & -                          & 0.930                   & 0.782                   & 0.849                        \\ \hline
\end{tabular}
\label{valid}
\end{table}

\subsection{Learning to Rank}
Learning-to-rank, a popular machine learning method, is widely applied in various information retrieval competitions~\cite{yang2022thuir,li2023towards,chen2023thuir}. To further improve the performance, we employ learning-to-rank techniques in legal case entailment task, trying to mine the intrinsic qualities of different features with the gradient boosting framework.

More specifically, we extract all the feature scores listed in Table \ref{feauture} and apply LightGBM to estimate the final scores of all query-candidate pairs. By fitting on the training set, we can learn the weights of different features. Finally, we choose the model that achieves the best NDCG@1 on the validation set for testing.

\section{Experiment Result}
The performance comparisons of different methods on the validation set are shown in Table \ref{valid}. Since the average number of relevant paragraphs is about one for each query, we evaluate the precision, recall, and f1-score at the cut-off value of 1.
From the experimental results, we have the following observations:

\begin{itemize}[leftmargin=*]
    \item As simple lexical matching cannot accurately capture the implication relationship between paragraphs, traditional methods such as BM25 and QLD have a more average performance.
    \item Benefit from supervised data, pre-trained language model further improves performance. The model with more parameters usually has a better performance. The monoT5-3B achieves the best results among all single models. This indicates that more parameters facilitate the understanding of the legal case entailment.
    \item Surprisingly, LEGAL-BERT-base with 110m parameters outperforms BERT-large and RoBERTa-large, which indicates that legal-oriented pre-training tasks allow the language model to have more legal knowledge and thus achieve better performance.
    \item Unexpectedly, the performance of FLAN-T5 drops dramatically. We assume that this is due to the inconsistency between the instruction fine-tuning process and the downstream tasks. In the future, we will explore more prompt formats to exploit the potential of large language models for legal case entailment task.
    \item Learning to rank techniques significantly improves the performance on the validation set. However, in the final leaderboard, the performance decreases after learning to rank on the contrary. We think this is due to overfitting due to a few training data. Also, it may indicate that the answer paragraphs cannot be simply confirmed by information retrieval techniques.
\end{itemize}

Overall, more parameters and more legal knowledge can help language models perform better on legal case entailment tasks. In the future, we will explore the application of large legal language models to legal case entailment task.

The final top5 results of COLIEE2023 task 2 are shown in Table \ref{test}. The run with monoT5 has the third placement and the run with ensemble placed fifth. When we get the test set labels, we retrain the learning-to-rank model and choose a smaller early stop step. The results on the test set are reported in the last row of Table \ref{test}. We can find that learning to rank techniques can slightly improve performance by avoiding overfitting.

\begin{table}[]
\caption{Final top-5 of COLIEE 2023 Task 2 on the test set.}
\begin{tabular}{ccccc}
\hline
\textbf{Team} & \textbf{Submission} & \textbf{Precision} & \textbf{Recall} & \textbf{F1} \\ \hline
CAPTAIN       & mt5l-ed                  & 0.7870             & 0.7083          & 0.7456               \\
CAPTAIN       & mt5l-ed4                 & 0.7864             & 0.6750          & 0.7265               \\
THUIR         & thuir-monot5             & 0.7900             & 0.6583          & 0.7182               \\
CAPTAIN       & mt5l-e2                  & 0.7596             & 0.6583          & 0.7054               \\
THUIR         & thuir-ensemble\_2        & 0.7315             & 0.6583          & 0.6930               \\ \hline
THUIR         & new ensemble             & 0.8000               & 0.6667          & 0.7273          \\ \hline
\end{tabular}
\label{test}
\end{table}

\section{Conclusion}
This paper shows our solution for the COLIEE2023 legal entailment task. We experiments with lexical matching model, cross encoder and sequence to sequence model. Learning to sort techniques is employed to get the final score. We finally achieve third place in this competition. Results show that more parameters and legal knowledge contribute to legal case entailment. In the future, we will design larger models with legal-oriented pre-training tasks.

\clearpage
\balance
\bibliographystyle{ACM-Reference-Format}
\bibliography{sample-base.bib}
\end{document}